\documentclass{article}

\usepackage{microtype}
\usepackage{graphicx}
\usepackage{subfigure}
\usepackage{booktabs} 

\usepackage{hyperref}
\usepackage{multirow}

\usepackage[noend]{algorithmic}

\usepackage[accepted]{icml2023}

\usepackage{amsmath}
\usepackage{amssymb}
\usepackage{mathtools}
\usepackage{amsthm}

\usepackage[capitalize,noabbrev]{cleveref}

\crefname{defn}{Definition}{Definition}
\crefname{section}{Section}{Section}
\crefname{algorithm}{Algorithm}{Algorithm} 
\crefname{thm}{Thm.}{Thm.}
\crefname{lem}{Lemma}{Lemma}
\crefname{prop}{Prop.}{Prop.}
\crefname{asm}{Asm.}{Asm.}
\crefname{appendix}{Appx.}{Appx.}
\crefname{equation}{Equation}{Equations}
\crefname{figure}{Figure}{Figure}
\crefname{table}{Table}{Table}
\crefname{cor}{Corollary}{Corollary}

\theoremstyle{plain}

\theoremstyle{definition}

\theoremstyle{remark}

\usepackage[textsize=tiny]{todonotes}

\usepackage{tikz}
\usetikzlibrary{tikzmark}

\usepackage{notation}
\usepackage{researchpack}

\icmltitlerunning{Understanding the Distillation Process from Deep Generative Models to Tractable Probabilistic Circuits}

\begin{document}

\twocolumn[
\icmltitle{
Understanding the Distillation Process \\ from Deep Generative Models to Tractable Probabilistic Circuits
}



\icmlsetsymbol{equal}{*}

\begin{icmlauthorlist}
\icmlauthor{Xuejie Liu}{equal,thu}
\icmlauthor{Anji Liu}{equal,ucla}
\icmlauthor{Guy Van den Broeck}{ucla}
\icmlauthor{Yitao Liang}{pku,bigai}
\end{icmlauthorlist}

\icmlaffiliation{pku}{Institute for Artificial Intelligence, Peking University, P.R. China}
\icmlaffiliation{thu}{Department of Automation, Tsinghua University, P.R. China}
\icmlaffiliation{ucla}{Computer Science Department, University of California, Los Angeles, USA}
\icmlaffiliation{bigai}{Beijing Institute for General Artificial Intelligence (BIGAI)}

\icmlcorrespondingauthor{Xuejie Liu}{liebenxj@gmail.com}

\icmlkeywords{Machine Learning, ICML}

\vskip 0.3in
]



\printAffiliationsAndNotice{\icmlEqualContribution} 

\begin{abstract}
Probabilistic Circuits (PCs) are a general and unified computational framework for tractable probabilistic models that support efficient computation of various inference tasks (e.g., computing marginal probabilities). Towards enabling such reasoning capabilities in complex real-world tasks, \citet{liu2022scaling} propose to distill knowledge (through latent variable assignments) from less tractable but more expressive deep generative models. However, it is still unclear what factors make this distillation work well. In this paper, we theoretically and empirically discover that the performance of a PC can exceed that of its teacher model. Therefore, instead of performing distillation from the most expressive deep generative model, we study \emph{what properties the teacher model and the PC should have in order to achieve good distillation performance}. This leads to a generic algorithmic improvement as well as other data-type-specific ones over the existing latent variable distillation pipeline. Empirically, we outperform SoTA TPMs by a large margin on challenging image modeling benchmarks. In particular, on ImageNet32, PCs achieve 4.06 bits-per-dimension, which is only 0.34 behind variational diffusion models \citep{kingma2021variational}.
\end{abstract}

\section{Introduction}
\label{sec:intro}



Developing Tractable Probabilistic Models (TPMs) that are capable of performing various inference tasks (\eg computing marginals) is of great importance as they enable a wide range of downstream applications such as constrained generation \citep{peharz2020einsum,correia2020joints}, causal inference \citep{wang2022symbolic}, and data compression \citep{liu2021lossless}. Probabilistic Circuits (PCs) \citep{choi2020probabilistic} refer to a class of TPMs with similar representations, including Sum-Product Networks \citep{poon2011sum}, and-or search spaces \citep{marinescu2005and}, and arithmetic circuits \citep{darwiche2002logical}. To take full advantage of the attractive inference properties of PCs, a key challenge is to improve their modeling performance on complex real-world datasets.

There have been significant recent efforts to scale up and improve PCs from both algorithmic \citep{correia2022continuous,shih2021hyperspns,dang2022sparse,peharz2020random} and architectural \citep{peharz2020einsum,dang2021juice} perspectives. In particular, \citet{liu2022scaling} propose the Latent Variable Distillation (LVD) pipeline that uses less-tractable yet more expressive Deep Generative Models (DGMs) to provide extra supervision to overcome the suboptimality of Expectation-Maximization (EM) based PC parameter learners. With LVD, PCs are able to achieve competitive performance against some widely used DGMs on challenging datasets such as ImageNet32 \citep{deng2009imagenet}.

\begin{figure}[t]
    \centering
    \includegraphics[width=\columnwidth]{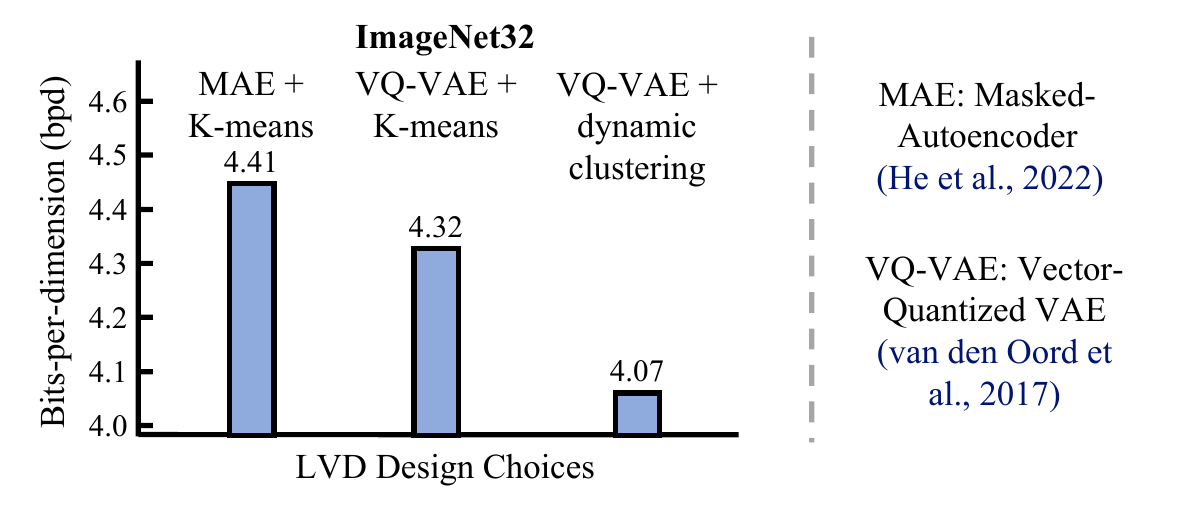}
    \vspace{-2.6em}
    \caption{Different design choices in the LVD pipeline lead to drastically different performance (lower is better) on ImageNet32. All LVD-learned PCs have $\sim\!\!200M$ parameters.}
    \label{fig:lvd-diffs}
    \vspace{-2.0em}
\end{figure}

However, despite its great potential, we have a limited understanding of when and how LVD leads to better modeling performance. As a result, the success of existing instantiations of the LVD pipeline relies heavily on trial and error. For example, as shown in \cref{fig:lvd-diffs}, modeling performance varies significantly as we change the DGM or the strategy to obtain supervision from them, even when the size of the PCs are similar.

This work aims to demystify the LVD pipeline and provide practical design guidelines for image data.
By interpreting LVD from a variational inference perspective, we show that the performance of LVD-learned PCs is not necessarily upper bounded by their teacher DGMs. This is in sharp contrast with distilling knowledge from a large neural network to a smaller one, where the training performance typically degrades \citep{gou2021knowledge}. Therefore, instead of trying to use the SoTA DGM to perform LVD, we should focus on a more fundamental question: \emph{what properties of the teacher DGM would lead to better performance of LVD-learned PCs?} Although there is still no definite answer, we identify practical design guidelines that lead to expressive yet compact PCs.

Following the guidelines, we observe a general deficit in the existing LVD pipeline. Specifically, due to the mismatch between the discrete latent variable assignments requested by PCs and the continuous neural representations, a one-shot discretization method is often used. However, this causes significant information loss and leads to degraded modeling performance. To overcome this problem, we propose a \emph{progressive growing} algorithm to leverage feedback from the PC to perform dynamic clustering, thus minimizing the performance loss caused by discretization. Progressive growing is also able to exploit reusable sub-structures, which leads to compact yet expressive PCs. Together with several image-specific design choices derived from the guidelines, we are able to out-perform SoTA TPMs by a large margin on three challenging image-modeling datasets: CIFAR \citep{krizhevsky2009learning} and two down-sampled ImageNet datasets \citep{deng2009imagenet}. In particular, we achieve 4.06 bits-per-dimension on ImageNet32, which is better than some intractable Flow models and VAEs such as Glow \citep{kingma2018glow} and only 0.34 less than the SoTA diffusion-based DGM \citep{kingma2021variational}.

\section{Background}
\label{sec:bg}

This section introduces PCs (Sec.~\ref{sec:bg-pc}) and the Latent Variable Distillation (LVD) pipeline (Sec.~\ref{sec:bg-lvd}).

\subsection{Probabilistic Circuits}
\label{sec:bg-pc}

Probabilistic circuits (PCs) are a broad class of TPMs that characterize probability distributions as deep computation graphs. The syntax and semantics of PCs are as follows.

\begin{defn}[Probabilistic Circuits] Represented as a  parameterized directed acyclic computation graph (DAG), a PC $p(\X)$ defines a joint distribution over a set of random variables $\X$ by a single root node $n_r$. The nodes in the DAG are divided into three types of computational units: \textit{input}, \textit{sum}, and \textit{product}. Notably, each leaf node in the DAG serves as an \textit{input} unit, while an inner node can be subdivided into a \textit{sum} unit or a \textit{product} unit according to its mechanism for combining child distributions. In the forward path, every inner node receives \textit{inputs} from its children (denoted $\ch(n)$) and computes \textit{outputs}, thus encoding a probability distribution $p_n$ in a recursive fashion:
\begin{align*}
    p_n(\boldsymbol{x}):= 
    \begin{cases}
        f_n(\boldsymbol{x}) & \text { if } n \text { is an input unit, } \\ 
        \sum_{c \in \operatorname{in}(n)} \theta_{n, c} \cdot p_c(\boldsymbol{x}) & \text { if } n \text { is a sum unit, } \\ 
        \prod_{c \in \operatorname{in}(n)} p_c(\boldsymbol{x}) & \text { if } n \text { is a product unit, }
    \end{cases}
\end{align*}

where $f_n(\boldsymbol{x})$ is a univariate probability distribution (\eg Gaussian, Categorical), and $\theta_{n,c}$ represents the parameter corresponding to edge $(n, c)$ in the DAG. Intuitively, a \textit{sum} unit models a weighted mixture of its children's distributions, which requires all its edge parameters to be non-negative and sum up to one, i.e., $\sum_{c\in \ch(n)} \theta_{n,c} = 1, \theta_{n,c} \geq 0$. And a product unit encodes a factorized distribution over its children. Finally, a PC represents the distribution encoded by its root node $n_r$. Additionally, we assume w.l.o.g. that a PC alternates between the sum and product layers before reaching its inputs.
\end{defn}

The ability to answer numerous probabilistic queries (\eg marginals, entropies, and divergences) \citep{vergari2021compositional} exactly and efficiently distinguishes PCs from various deep generative models. Such ability is typically interpreted as \textit{tractability}. To guarantee PCs' tractability, certain structural constraints have to be imposed on their DAG structure. For instance, \textit{smoothness} together with \textit{decomposability} ensure that a PC can compute arbitrary marginal probabilities in linear time \wrt its size, which is the number of edges in its DAG. These are properties of the variable scope $\phi(n)$ of PC unit $n$, that is, the variable set comprising all its descendent input nodes.

\begin{defn}[Decomposability]
A PC is decomposable if for every product unit $n$, its children have
disjoint scopes: 
    \begin{align*}
        \forall c_1, c_2 \in \ch(n) \, (c_1 \neq c_2), \; \phi(c_1) \cap \phi(c_2) = \varnothing.
    \end{align*}
\end{defn}

\begin{defn}[Smoothness] 
A PC is smooth if for every sum unit $n$, its children have the same scope: 
    \begin{align*}
        \forall c_1, c_2 \in \ch(n), \; \phi(c_1) = \phi(c_2).
    \end{align*}
\end{defn}

\subsection{Latent Variable Distillation}
\label{sec:bg-lvd}


Despite the recent breakthroughs in developing efficient computational frameworks for PCs \citep{dang2021juice,molina2019spflow}, exploiting the additional expressive power of large-scale PCs remains extremely challenging. Abundant empirical evidence has attributed this phenomenon to the failure of existing EM-based optimizers to find good local optima in the large and hierarchically nested latent space of PCs \citep{peharz2016latent}, which is defined by the hierarchically distributed sum units in their DAGs


Latent Variable Distillation (LVD) overcomes the aforementioned bottleneck by providing extra supervision to PC optimizers through semantic-aware latent variable (LV) assignments, which are acquired from less tractable yet more expressive deep generative models \citep{liu2022scaling}. Specifically, LVD operates by first materializing some/all LVs in the PC. That is, transforming the original PC $\p(\X)$ into $\p(\X, \Z)$ whose marginal distribution over $\X$ stays unchanged, \ie $\p(\X) = \sum_{\z} \p(\X, \Z \!=\! \z)$.


Next, deep generative models (DGMs) are used to induce semantic-aware assignments of LVs $\Z$ for every training sample $\x \!\in\! \data_{\mathrm{train}}$, leading to an augmented dataset $\data_{\mathrm{aug}} \!:=\! \{ (\x, \z) : \x \!\in\! \data_{\mathrm{train}} \}$. This LV induction step can be done in various ways and with different DGMs. For example, in \citet{liu2022scaling}, $\Z$ is obtained by clustering the latent features produced by a Masked Autoencoder \citep{he2022masked}. 


Finally, the augmented dataset $\data_{\mathrm{aug}}$ is used to maximize a lower bound of the log-likelihood, as shown on the right-most term:
    {\setlength{\abovedisplayskip}{0.4em}
    \setlength{\belowdisplayskip}{0.2em}
    \begin{equation}
    \begin{aligned}
        \sum_{i=1}^N \log p\left(\boldsymbol{x}^{(i)}\right) & := \sum_{i=1}^N \log \sum_{\boldsymbol{z}} p\left(\boldsymbol{x}^{(i)}, \boldsymbol{z}\right), \\ 
        & \geq \sum_{i=1}^N \log p\left(\boldsymbol{x}^{(i)}, \boldsymbol{z}^{(i)}\right).
    \label{eq:lvd-objective}
    \end{aligned}
    \end{equation}}

After training with the augmented dataset, we can obtain the target distribution $\p(\X)$ by marginalizing out $\Z$, which can be done in linear (\wrt size of the PC) time \citep{choi2020probabilistic}. The PC can then be finetuned with the original dataset to improve performance further.

The success of LVD is primarily attributed to its ability to simplify the size and depth of PCs' deeply nested latent variable spaces \citep{peharz2016latent}. Specifically, after LV materialization, supervision of the LVs can be provided by DGMs, and EM-based PC parameter learners are only responsible for inferring the values of the remaining implicitly defined LVs. Since the DGMs guide PC learning through their provided LV assignments, we refer to them as teacher models and the PCs as student models.



\section{Characterizing Performance Gaps in LVD}
\label{sec:problem}


Although LVD has demonstrated its potential to boost the performance of large PCs, its effectiveness depends strongly on the design choice of materialized LVs and how they are induced from external sources. Specifically, as shown in \cref{fig:lvd-diffs}, a bad design choice will lead to significantly worse performance, while a good one can further close the performance gap with SoTA intractable DGMs. 
Therefore, a crucial yet unanswered question concerning LVD is: \emph{what are the design principles for the LV induction process to achieve good modeling performance?}



We provide a preliminary answer to this question by characterizing the performance differences between the teacher DGM and the student PC via variational inference (VI), which is the mathematical foundation of various DGMs such as VAEs \citep{kingma2013auto} and Diffusion models \citep{ho2020denoising}. Consider a latent variable model $\p_{\theta}(\x) := \sum_{\z} \p_{\theta}(\x \given \z) \p_{\theta}(\z)$. Instead of directly maximizing the log-likelihood $\log \p_{\theta}(\x)$, which could be infeasible, VI proposes to also learn a variational posterior $\q_{\phi}(\z \given \x)$ and maximize the following evidence lower bound (ELBO) of the log-likelihood:
    \begin{align}
        \mathbb{E}_{\z \sim q_\phi(\cdot \given \x)} \left [ \log p_\theta(\x \given \z) \right ] -D_{\mathrm{KL}}\left(q_\phi(\z \given \x) \| p_\theta(\z)\right). \label{eq:elbo}
    \end{align}
Consider a PC $\p_{\mathrm{pc}}(\x) := \sum_{\z} \p_{\mathrm{pc}} (\x \given \z) \p_{\mathrm{pc}} (\z)$ defined on the same $\X$ and $\Z$ as above. The ultimate goal of LVD is to distill knowledge from $\p_{\theta}(\x)$ to $\p_{\mathrm{pc}}(\x)$ to maximize the PC's log-likelihood.

\begin{figure}
    \centering
    \includegraphics[width=\columnwidth]{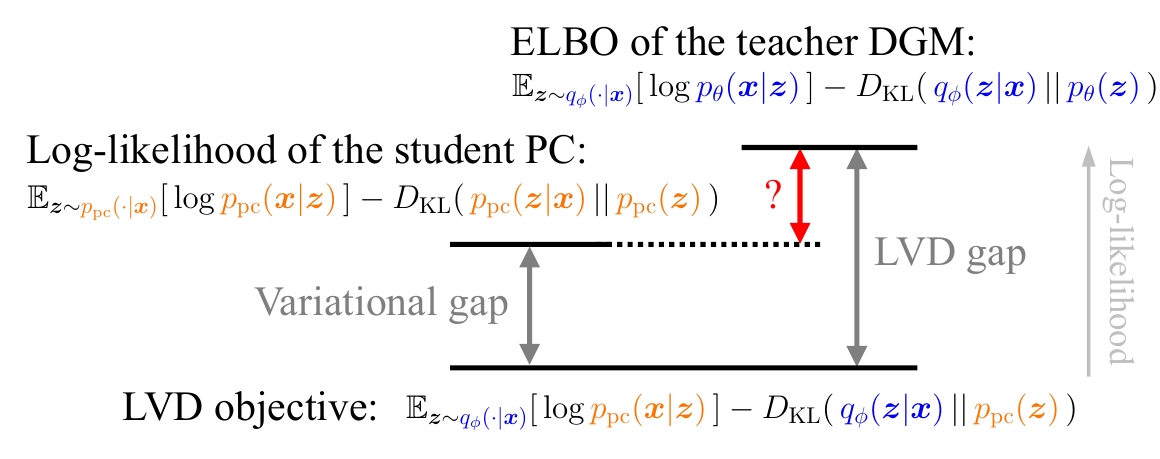}
    \vspace{-2.4em}
    \caption{Performance difference of the teacher DGM (top-right) and the student PC (top-left) is characterized by the relative significance between the variational gap and the LVD gap.}
    \label{fig:elbo-gap}
\end{figure}


\begin{figure*}[t]
    \centering
    \begin{minipage}[c]{0.30\textwidth}
        \includegraphics[width=\textwidth]{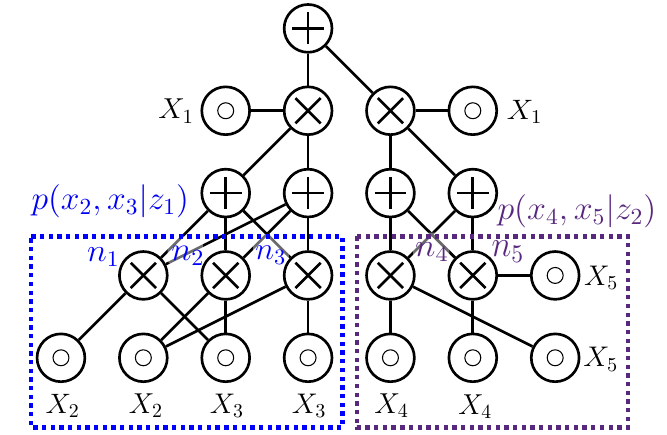}
        \vspace{-2.4em}
        \caption{Example cluster-conditioned distributions represented by a PC.}
        \label{fig:cls-cond-pc}
        \includegraphics[width=\textwidth]{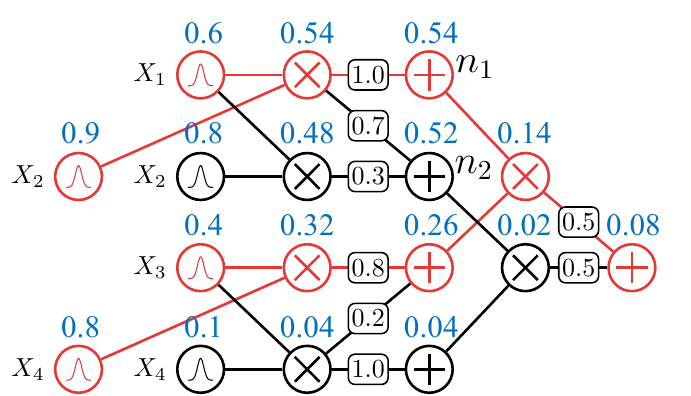}
        \vspace{-2.4em}
        \caption{An example PC with likelihood of every node \wrt $\x$ labeled blue. Red nodes are ``important'' for modeling $\x$.}
        \label{fig:flow-illustration}
    \end{minipage}
    \hfill
    \begin{minipage}[c]{0.68\textwidth}
        \includegraphics[width=\textwidth]{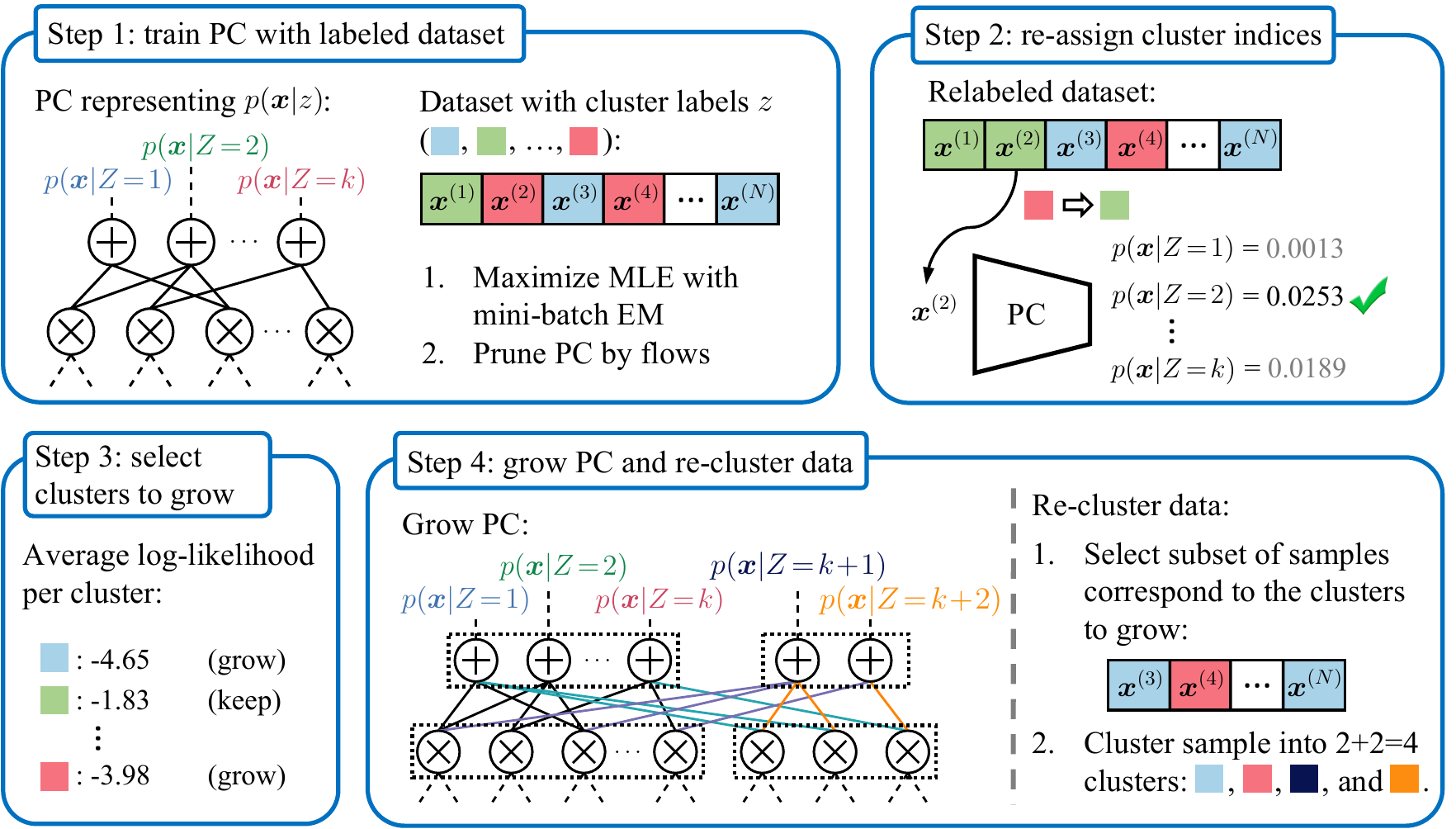}
        \vspace{-2.2em}
        \caption{Illustration of the proposed progressive growing algorithm. The algorithm takes as input a dataset of pairs $(\x, \h)$, where $\h = g_{\phi}(\x)$ is a continuous neural representation of $\x$, and a PC representing $\p(\x \given z)$. The algorithm iterates through the four steps shown above to gradually expand cluster size (\ie $\abs{\Z}$) and grow the PC.}
        \label{fig:prog-grow}
    \end{minipage}
    \vspace{-0.4em}
\end{figure*}

A natural way to achieve this is to use $\q_{\phi}(\z \given \x)$ as the variational posterior for the PC. This leads to the following ELBO objective:
    \begin{align}
        \mathbb{E}_{\z \sim \q_{\phi}(\cdot \given \x)\!} \left [ \log p_{\mathrm{pc}}(\x\given\z) \right ] \!-\! D_{\mathrm{KL}}\! \left(\q_{\phi}(\z \given \x) \| p_{\mathrm{pc}}(\z) \right ). \label{eq:lvd-vi}
    \end{align}
Although written in different forms, this ELBO objective is equivalent to \cref{eq:lvd-objective} up to a constant factor independent of the PC, (see Appx.~\ref{appx:equiv-lvd} for a rigorous elaboration). Intuitively, $\q_{\phi} (\z \given \x)$ is treated as the external model to induce LV assignments $\z$ for every training sample $\x$. Therefore, we call \cref{eq:lvd-vi} the LVD objective.

The LVD objective provides a bridge to characterize the difference between the performance of the teacher DGM (Eq.~(\ref{eq:elbo})) and the log-likelihood of the student PC. Specifically, as shown in the \cref{fig:elbo-gap}, the performance gap between the teacher DGM and the LVD objective, termed the LVD gap, characterizes the performance loss of LVD. However, the final performance difference between the teacher DGM and the student PC can be much less than the LVD gap. Specifically, thanks to the tractability of PC, $\p_{\mathrm{pc}} (\z \given \x)$ can be obtained in closed form. Therefore, the variational gap between the LVD objective and the PC's log-likelihood can be closed ``for free'' right after $\p_{\mathrm{pc}} (\x \given \z)$ and $\p_{\mathrm{pc}} (\z)$ are trained by the LVD objective. That is, as demonstrated in \cref{fig:elbo-gap}, after LVD training, which gives the ELBO shown at the bottom, we can directly obtain a PC with log-likelihood shown at the top-left side.

Perhaps surprisingly, the above analysis suggests that the log-likelihood of the student PC is not necessarily upper bounded by the ELBO of the teacher DGM. Specifically, as illustrated in \cref{fig:elbo-gap}, this happens whenever the variational gap is larger than the LVD gap. In the extreme case where the student perfectly simulates the teacher (\ie $\p_{\theta} (\x \given \z) \!=\! \p_{\mathrm{pc}} (\x \given \z)$ and $\p_{\theta} (\z) \!=\! \p_{\mathrm{pc}} (\z)$), the PC becomes a tractable instantiation of $\p_{\theta} (\x) = \sum_{\z} \p_{\theta} (\x \given \z) \p_{\theta} (\z)$ that can compute various probabilistic queries. Towards achieving this ideal case, we need to minimize the LVD gap.

A key insight towards closing the LVD gap is to ensure $\p_{\theta} (\x \given \z)$ (resp. $\p_{\theta} (\z)$) has similar modeling assumptions to $\p_{\mathrm{pc}} (\x \given \z)$ (resp. $\p_{\mathrm{pc}} (\z)$). This works well in both directions: on the one hand, by mimicking the inductive biases of the DGM (\ie $\p_{\theta} (\x \given \z)$ and $\p_{\theta} (\z)$), we can learn PCs that have better performance as well as fewer parameters; on the other hand, it is often beneficial to remove modeling assumptions in the DGM that cannot be fully adopted by PCs due to their structural constraints, though it might lead to worse performance of the DGM. In the following, we first identify a general source of modeling assumption mismatch, and propose an algorithm to mitigate this problem (Sec.~\ref{sec:progressive-grow}). We then demonstrate how these design principles specialize to image data and improve modeling performance (Sec.~\ref{sec:image-modeling}).

\section{Latent Variable Distillation from Continuous Neural Representations}
\label{sec:progressive-grow}


A major challenge in minimizing the LVD gap is the mismatch between expressive \emph{continuous} neural embeddings and the \emph{discrete} nature of LVs materialized from PCs (since sum units represent discrete mixtures). Therefore, to obtain discrete LV assignments, either a post hoc discretization step (\eg K-means) is used, or the DGM needs to learn discrete representations, which often leads to worse performance compared to learning continuous features. Such one-shot discretization strategies result in a relatively large LVD gap, which degrades PC modeling performance significantly. 

Formally, the variational posterior of the DGM can be decomposed as 
    \begin{align*}
        \q_{\phi} (\z \given \x) := \q (\z \given \h), \text{~where~} \h = g_{\phi} (\x).
    \end{align*}
Here $g_{\phi}$ is a neural network, and $\h$ is a continuous neural representation. Although we still have to discretize $\h$ to obtain LV assignments for the PC, this discretization procedure need not be one-shot: we can leverage \emph{feedback} from the PC to adjust and re-assign cluster indices to narrow the LVD gap. Specifically, we want the cluster indices (\ie $\z$) to be assigned in a way that both respect the neural representation $\h$ (\ie samples with similar $\h$ are assigned to the same cluster) and are easily learned by the cluster-conditioned PCs $\p_{\mathrm{pc}} (\x \given \z)$. Intuitively, while the latter condition ensures $\p_{\mathrm{pc}} (\x \given \z)$ is properly learned, the former guarantees that $\z$ preserves information from $\h$, which empirically leads to a better $\p_{\mathrm{pc}} (\z)$.

Before delving into the details of our solution, we briefly review the LV materialization process and illustrate the structure of cluster-conditioned PCs $\p(\x \given \z)$. Instead of assigning every sum unit an LV (as they represent mixture distributions), we group them according to their variable scopes (\cf Sec.~\ref{sec:bg-pc}), and assign every LV to a particular scope. Specifically, since the children of the sum units with every scope $\phi$ are all product units with the same scope, we can assign each child product unit a unique discrete value. Take the PC in \cref{fig:cls-cond-pc} as an example, we choose to materialize two LVs $Z_1$ and $Z_2$ \wrt the scopes $\phi_1 \!:=\! \{X_2,X_3\}$ and $\phi_2 \!:=\! \{X_4,X_5\}$, respectively. After LV materialization, we obtain two sub-PCs representing the cluster-conditioned distributions $p(x_2, x_3 \given z_1)$ and $p(x_4, x_5 \given z_2)$, respectively. Specifically, the child product node $n_i$ ($\forall i \!\in\! \{1,2,3\}$) with scope $\phi_1$ represents $p(x_2,x_3\given Z_1 \!=\! i)$. Therefore, $\p(x_2, x_3 \given z_1)$ is represented by the three-headed PC highlighted in the dashed blue box. 


We proceed to describe the proposed \emph{progressive growing} algorithm that overcomes the suboptimality of the aforementioned one-shot discretization method. The algorithm takes as input a dataset $\data_{\mathrm{train}}$ accompanied with continuous neural embeddings, defined as $\data \!:=\! \{(\x, \h) \!:\! \x \!\in\! \data_{\mathrm{train}}, \h \!=\! g_{\phi}(\x)\}$. Having materialized a LV $Z$ that corresponds to scope $\X$, the algorithm also takes an initial cluster-conditioned PC $\p(\x \given z)$ as input. We assume $Z$ initially takes a single value (\ie all samples in $\data_{\mathrm{train}}$ belong to the same cluster), and thus $\p(\x \given z)$ is represented by a single-headed PC.

Given a predefined number of clusters, denoted $K$, progressive growing aims to learn both a discretization function that maps every $\h$ into a cluster index $i \in [K]$, and a $K$-headed PC representing $\p(\x \given Z \!=\! i)$ ($\forall i \!\in\! [K]$). This is done by iteratively dividing $\data$ into more clusters and correspondingly learning the structure and parameters of the cluster-conditioned PCs. Specifically, as illustrated in \cref{fig:prog-grow}, progressive growing operates by repeating four main steps, which are detailed in the following.

\textbf{Step 1: Training PC with Labeled Dataset.}\quad In this stage, we have access to a clustering function $\lambda_{k}$ that maps every $\h$ to an index in $[k]$, where $1 \!\leq\! k \!\leq\! K$ is the current number of clusters, and a $k$-headed PC with the $i$th head encoding $\p(\x \given Z=i)$. We train the PC by maximizing the conditional log-likelihood specified by $\data$ and $\lambda_{m}$:
    \begin{align}
        \maximize_{\boldsymbol{\varphi}} \sum_{(\x, \h) \in \data} \log \p_{\boldsymbol{\varphi}} (\x \given Z = \lambda_{k} (\h)), \label{eq:pg-mle}
    \end{align}
\noindent where $\boldsymbol{\varphi}$ are the parameters of the PC. We optimize \cref{eq:pg-mle} with the standard mini-batch EM algorithm \citep{peharz2020einsum,choi2021group}. Hyperparameters are detailed in \cref{appx:prog-grow}. To learn a compact yet expressive PC, we apply the pruning algorithm proposed by \citet{dang2022sparse} after the parameter learning phase. This results in significantly smaller PCs with negligible performance loss.

\textbf{Step 2: Re-assigning Cluster Indices.}\quad As hinted by the suboptimality of the one-step discretization method, cluster indices assigned by $\lambda_{k}$ may not fully respect the PC $\p(\x \given z)$. That is, since $\lambda_k$ is obtained by clustering neural representation $\h$, some samples $\x$ assigned to cluster $i$ could be better modeled by $\p_{\boldsymbol{\varphi}} (\x \given Z \!=\! j)$ ($j \neq i$) trained in the previous step. To mitigate this problem, we leverage feedback from the PC to re-assign cluster indices. Specifically, as demonstrated in \cref{fig:prog-grow}, the cluster index of sample $\x$ is re-labeled as $z := \argmax_{i \in [k]} \p(\x \given Z = i)$. Function $\lambda_{k}$ is modified correspondingly to reflect this change.

As we will elaborate more in the following steps, this re-labeling process allows us to escape from poorly assigned clusters in past iterations, and is crucial to the effectiveness of progressive growing.


\textbf{Step 3: Selecting Clusters to Grow.}\quad As suggested by its name, progressive growing operates by iteratively expanding the number of clusters in $Z$. To improve the overall performance of the cluster-conditioned PC (\ie Eq.~\ref{eq:pg-mle}), we select clusters with low average log-likelihood to be further divided. Specifically, as illustrated in \cref{fig:prog-grow}, we first compute the average log-likelihood for each cluster $i \in [k]$: 
    \begin{align*}
        \mathtt{LL}_i := \frac{1}{\abs{\data_i}} \sum_{\x \in \data_i} \log \p (\x \given Z = i),
    \end{align*}
\noindent where $\data_i \!:=\! \{\x \!:\! (\x, \h) \!\in\! \data, \lambda_k (\h) \!=\! i\}$. We then select a subset of clusters based on $\{\mathtt{LL}_i\}_{i=1}^{k}$ and the number of samples belonging to every cluster. See \cref{appx:prog-grow} for detailed design choices.

\textbf{Step 4: Growing PC and Re-clustering Data.}\quad Suppose the previous step selects a set of cluster indices $\calI$ for growing. The goal of this step is to expand these $\abs{\calI}$ clusters into $M$ new clusters ($M > \abs{\calI}$). Under the hood, we need to re-cluster the corresponding subset of samples as well as apply structure modifications to the PC to fit the new clusters. Both procedures are described in the following.

To ensure that the structure and parameters of the multi-headed PC are still relevant to the cluster assignments $\lambda_k$ after the reclustering step, a natural approach is to perform hierarchical growing and clustering to the PC and the dataset, respectively. Specifically, for each selected cluster $i \!\in\! \calI$, we use K-means to cluster the samples belonging to the $i$th cluster into $n$ clusters, and create $n\!-\!1$ new PC root units for the added clusters based on $\p_{\boldsymbol{\varphi}} (\x \given Z \!=\! i)$. We use a slightly modified approach to re-cluster training samples for all $\abs{\calI}$ clusters simultaneously. Specifically, we first select the subset of samples $(\x, \h)$ belonging to clusters in $\calI$. We then run K-means to cluster the neural representations $\h$ into $K$ clusters, with the first $\abs{\calI}$ cluster centers initialized to be the centers of the clusters in $\calI$. $\lambda_k$ is then updated to reflect the new cluster assignments. In this way, the first $\abs{\calI}$ new clusters are still relevant to the corresponding PC $\p_{\boldsymbol{\varphi}} (\x \given Z \!=\! i)$ ($i \!\in\! \calI$).


In order to represent the newly-added clusters, the structure of the PC needs to be modified to contain $M - \abs{\calI}$ additional root/head units to represent $\p(\x \given Z = i)$ ($i \in \{k+1, \dots, k+M-\abs{\calI}\}$). A simple strategy would be to directly copy all descendent units of $M - \abs{\calI}$ existing root units for the new clusters. However, this will significantly increase the size of the cluster-conditioned PC, rendering the progressive growing algorithm highly inefficient. Moreover, it rules out the possibility to reuse sub-circuits that are useful for modeling $\x$ conditioned on different $z$, seriously limiting the PC's expressive power at any particular size.

To mitigate this problem, we propose a structure growing operator that only copies the most important substructure for describing a distribution. By introducing additional edges between the original and copied sub-circuit, the PC can learn to share structures that can be used to describe $\p(\x \given z)$ for various $z$. At the heart of the growing algorithm is a statistic termed \emph{flow} that measures the generative significance of a node/edge \wrt a sample $\x$ \citep{dang2022sparse,liu2021tractable}, defined as follows.

\begin{defn}[Circuit flow]
\label{def:flow}
For a PC $\p (\X)$ and a sample $\x$, the circuit flow for every PC unit $n$, denoted $F_{n}(\x)$, is defined recursively as follows ($\pa (n)$ denotes the set of parent units of $n$): first, $F_n(\x) = 1$ if $n$ is the root unit; next, if $n$ is a product unit, we have 
    {\setlength{\abovedisplayskip}{0.3em}
    \setlength{\belowdisplayskip}{0.3em}
    \begin{align*}
        F_{n}(\x) := \sum\nolimits_{m \in \pa(n)} \frac{\theta_{m,n} \cdot \p_{n}(\x)}{\p_{m}(\x)} \cdot F_m(\x);
    \end{align*}}
otherwise ($n$ is a sum or input unit), the flow is defined by 
    {\setlength{\abovedisplayskip}{0.3em}
    \setlength{\belowdisplayskip}{0.0em}
    \begin{align*}
        F_{n}(\x) := \sum\nolimits_{m \in \pa(n)} F_{m}(\x).
    \end{align*}}
\end{defn}

\begin{figure}[t]
\vspace{-0.4em}
\begin{algorithm}[H]
\caption{Grow Multi-Headed PCs}
\label{alg:grow}
{\fontsize{9}{9} \selectfont
\begin{algorithmic}[1]

\STATE {\bfseries Input:} A dataset $\data \!=\! \{(\x^{(i)}\!,\! z^{(i)})\}_{i=1}^{N}$, where $z^{(i)} \!\!\in\! [k]$ is the

$\qquad \; \; \;$ cluster index of $\x^{(i)}$; a $k$-headed PC $\p$

\STATE {\bfseries Output:} A new multi-headed PC $\p'$

\vspace{0.25em}

\STATE Compute $F_{n} (\data)$ for every PC unit $n$

\vspace{0.2em}

\STATE $\mathtt{G} \leftarrow \{n : F_n(\data) \geq \epsilon\}$, where $\epsilon$ is a predefined threshold

\vspace{0.2em}

\STATE $\mathtt{old2new} \leftarrow \mathtt{dict}()$ \hfill \textcolor[RGB]{115,119,123}{$\triangleright$ Maps $n$ to a pair of (new) nodes}

\vspace{0.2em}

\ForEach
\NoDo
\FOR{\tikzmarknode{a1}{} $\!\!$ $n$ traversed in postorder \textbf{do} \hfill \textcolor[RGB]{115,119,123}{$\triangleright$ Child before parent}}

\vspace{0.2em}

\STATE $\! \mathtt{ch}_1,\! \mathtt{ch}_2 \!\leftarrow\! \{\!\mathtt{old2new}[c][0]\}_{c \in \ch(n)},\! \{\!\mathtt{old2new}[c][1]\}_{c \in \ch(n)}$

\vspace{0.2em}

\STATE{\tikzmarknode{a2}{} \textbf{if} $n$ \textbf{isa} input unit \textbf{then}}

\vspace{0.2em}

\STATE \hspace{1em} $\mathtt{old2new}[n] \leftarrow (n, \mathtt{copy}(n)) \text{~if~} n \in \mathtt{G} \text{~else~} (n, n)$ 

\vspace{0.2em}

\STATE{\tikzmarknode{a3}{} \textbf{elif} $n$ \textbf{isa} product unit \textbf{then}}

\vspace{0.2em}

\STATE \hspace{1.0em} $\mathtt{old2new}[n] \leftarrow (\bigotimes(\mathtt{ch}_1), \bigotimes(\mathtt{ch}_2))$

\vspace{0.2em}

\STATE{\tikzmarknode{a4}{} \textbf{elif} $n \in \mathtt{G}$ \textbf{isa} sum unit \textbf{then}}

\vspace{0.2em}

\STATE \hspace{1.0em} $\mathtt{old2new}[n] \leftarrow (\bigoplus(\mathtt{ch}_1, \mathtt{ch}_2), \bigoplus(\mathtt{ch}_1, \mathtt{ch}_2))$

\vspace{0.2em}

\STATE{\tikzmarknode{a5}{} \textbf{elif} $n \not\in \mathtt{G}$ \textbf{isa} sum unit \textbf{then}}

\vspace{0.2em}

\STATE \hspace{1.0em} $\mathtt{old2new}[n] \leftarrow (\bigoplus(\mathtt{ch}_1, \mathtt{ch}_2), \mathtt{None})$

\vspace{0.2em}

\ENDFOR
\ForOnly
\ReDo

\STATE \textbf{return} A multi-head PC with root nodes 

$\qquad \; \; \; \; \{ \mathtt{old2new}[n] : n \text{~is~a~root~node~in~} \p \}$

\end{algorithmic}}
\end{algorithm}
\begin{tikzpicture}[overlay,remember picture]
    \draw[black,line width=0.6pt] ([xshift=-28pt,yshift=-3pt]a1.west) -- ([xshift=-28pt,yshift=-102pt]a1.west) -- ([xshift=-24pt,yshift=-102pt]a1.west);
    \draw[black,line width=0.6pt] ([xshift=4pt,yshift=-3pt]a2.west) -- ([xshift=4pt,yshift=-12pt]a2.west) -- ([xshift=8pt,yshift=-12pt]a2.west);
    \draw[black,line width=0.6pt] ([xshift=4pt,yshift=-3pt]a3.west) -- ([xshift=4pt,yshift=-12pt]a3.west) -- ([xshift=8pt,yshift=-12pt]a3.west);
    \draw[black,line width=0.6pt] ([xshift=4pt,yshift=-3pt]a4.west) -- ([xshift=4pt,yshift=-12pt]a4.west) -- ([xshift=8pt,yshift=-12pt]a4.west);
    \draw[black,line width=0.6pt] ([xshift=4pt,yshift=-3pt]a5.west) -- ([xshift=4pt,yshift=-12pt]a5.west) -- ([xshift=8pt,yshift=-12pt]a5.west);
\end{tikzpicture}
\vspace{-2.4em}
\end{figure}

Intuitively, flow $F_n (\x)$ quantifies the ``contribution'' of unit $n$ to the log-likelihood of $\x$. \cref{fig:flow-illustration} demonstrates an example PC-sample pair with likelihoods labeled on top of every node. Nodes and edges with relatively high flows are labeled red. Note that high node likelihood does not guarantee high flow, which is illustrated by $n_1$ and $n_2$: they both have high likelihoods, but only $n_1$ has high flow. For a dataset $\data$, $F_{n} (\data) := \sum_{\x \in \data} F_{n} (\x)$ measures the total contribution of $n$ to the samples in $\data$. 


Recall that our goal is to expand the current $k$-headed PC to have $M - \abs{\calI}$ additional root units to encode $\p(\x \given Z = i)$ ($i \in \{k \!+\! 1, \dots, k \!+\! M \!-\! \abs{\calI}\}$), respectively. To achieve this, we first extend \cref{def:flow} for multi-headed PC. Specifically, while the recurrent definition of the inner nodes remain unchanged, for the $i$th root node, we set the flow to $1$ if $\x$ is assigned to cluster $i$ by $\lambda_m$ and $0$ otherwise.

The proposed growing operator is shown in \cref{alg:grow}. It consists of two main parts: in lines 3-4, circuit flow is used to choose a subset of ``important'' (\ie nodes with flow higher than a predefined threshold) nodes to be grown; in lines 5-15, the PC is modified in a way that only the selected nodes are duplicated, while other parts are kept unchanged. In our use case, since we want to modify the sub-circuit corresponds to the $\abs{\calI}$ chosen clusters, we invoke \cref{alg:grow} with the subset of samples whose cluster indices are in $\calI$. According to the definition of flows, the returned PC will have $k + \abs{\calI}$ heads since the $\abs{\calI}$ chosen root nodes will be duplicated by the algorithm, while all other nodes will not. 

Progressive growing alternates between the four steps described above until we have expanded the number of clusters to a predefined value $K$. Therefore, the parameters of the multi-headed PC grown by step \#4 will be updated in step \#1 of the algorithm's next iteration.

In summary, the data re-clustering process in step 4 ensures that the cluster assignments respect the neural representation, and the cluster assigning process in step 2 leads to well-fitted cluster-conditioned PCs.

\section{Closing the LVD Gap for Image Data}
\label{sec:image-modeling}

Using image data as an example, this section demonstrates how the general guidelines for narrowing the LVD gap introduced in \cref{sec:problem} can be specialized to practical design choices. Throughout this paper, we adopt Vector Quantized Variational Autoencoders (VQ-VAEs) \citep{van2017neural,razavi2019generating} as the teacher model. In the following, we first briefly introduce VQ-VAE. We then proceed to describe the design choices we make to better align the modeling assumptions of $\p_{\theta} (\x \given \z)$ (resp. $\p_{\theta} (\z)$) and $\p_{\mathrm{pc}} (\x \given \z)$ (resp. $\p_{\mathrm{pc}} (\z)$).

\begin{figure}
    \centering
    \includegraphics[width=\columnwidth]{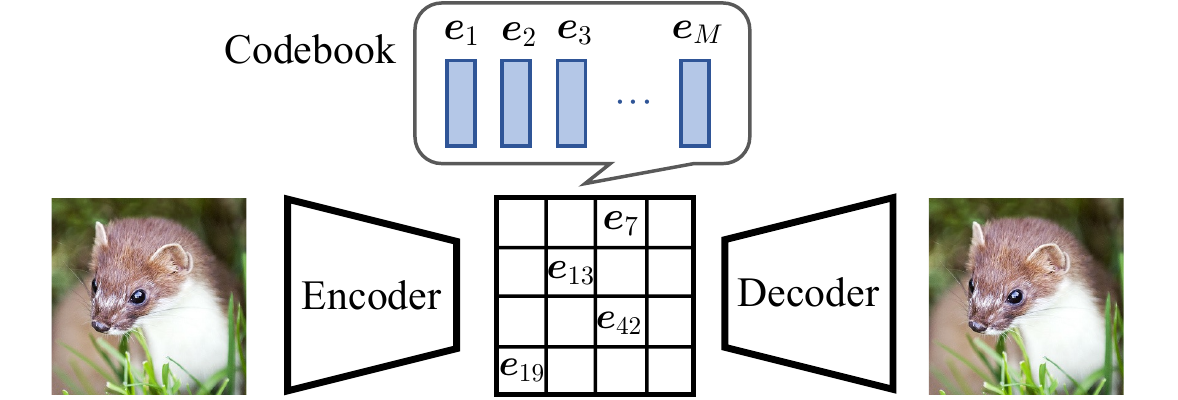}
    \vspace{-1.6em}
    \caption{Illustration of the VQ-VAE model.}
    \label{fig:vqvae}
    \vspace{-1.4em}
\end{figure}

\begin{table*}[t]
\caption{Density estimation performance of Tractable Probabilistic Models (TPMs) and Deep Generative Models (DGMs) on three natural image datasets. Reported numbers are test set bit-per-dimension (bpd), specifically, we report mean bpds and respective standard deviations over three runs. Bold indicates the best bpd (smaller is better) among all three TPMs. Best TPM w/o LVD represents the best performance over three TPMs: HCLT, RAT-SPN, and EiNet.}
\label{tab:img-main-results}
\vspace{0.2em}
\centering
\scalebox{1.0}{
\begin{tabular}{lccccccc}
    \toprule
    \multirow{2}{*}[-0.2em]{Dataset} & \multicolumn{3}{c}{TPMs} & \multicolumn{4}{c}{DGMs} \\
    \cmidrule(lr){2-4}
    \cmidrule(lr){5-8}
    ~ & LVD-PG (ours) & LVD & Best TPM w/o LVD & Glow & RealNVP & BIVA & VDM \\
    \midrule
    ImageNet32 & \textbf{4.06}{\scriptsize ±0.01} & 4.38 & 4.82 & 4.09 & 4.28 & 3.96 &  3.72\\
    ImageNet64 & \textbf{3.80}{\scriptsize ±0.07} & 4.12 & 4.67 & 3.81 & 3.98 & - & 3.40\\
    CIFAR & \textbf{3.87}{\scriptsize ±0.00} & 4.37 & 4.61 & 3.35 & 3.49 & 3.08 & 2.65 \\
    \bottomrule
\end{tabular}
}
\vspace{-1.0em}
\end{table*}

As shown in \cref{fig:vqvae}, VQ-VAE consists of an encoder that produces a feature map, and a decoder that reconstructs the input image using the feature map. Different from many other DGMs, the latent feature map of VQ-VAEs is constructed by a codebook with $M$ vectors representing $M$ codes. Specifically, the latent embedding at each position must be a vector from the codebook. Since every latent code in the feature map corresponds to a patch of the input image, we materialize an LV $Z_i$ for each position in the latent feature map, and define the corresponding image patch as $\X_i$. Denote $\Z := \{Z_i\}_i$, likelihood of an image $\x$ can be computed as $\p(\x) = \sum_{\z} \p(\z) \prod_{i} \p (\x_i \given z_i)$. For every $Z_i$, we can learn $\p(\x_i \given z_i)$ from the patches $\x_i$ and the corresponding continuous feature vectors $\h_i$ produced by VQ-VAE using the progressive growing algorithm detailed in the previous section. The generated discretization function can then be used to generate $\z = \{\z_i\}_{i}$ for every training sample, and is used to train $\p(\z)$.

However, in the above treatment, there are mismatches between the modeling assumptions made by VQ-VAE and the PC. First, since the latent feature map produced by VQ-VAE uses the same codebook at all locations, the cluster-conditioned distributions for different patches should be homogeneous. That is, for every discretization function $\lambda$ and sample pair $(\x, \h)$, we have $\forall i, j \in [\abs{\Z}]$,
    \begin{align*}
        \p(\X_i = \x \given Z_i = \lambda(\h)) \approx \p(\X_j = \x \given Z_j = \lambda(\h)).
    \end{align*}
To reflect this inductive bias, instead of learning $\p(\x \given z_i)$ for every $i \in [\abs{\Z}]$ independently, we aggregate their respective training samples and learn a single cluster-conditioned distribution, which is then applied to every image patch. That is, we do parameter tying between different cluster-conditioned distributions. This not only decreases the number of parameters of the PC, but also allows us to use much more data to train better cluster-conditioned distributions.

Another modeling assumption mismatch comes from the conditional independence between $\X_i$ and $\X \backslash \X_i$ given $Z_i$ assumed by the PC. The convolutional decoder of a VQ-VAE breaks this assumption as $\x_i$ can correlate to other patches given $\z_i$. To mitigate this mismatch, we use an independent decoder where $z_i$ is the only source of information used to generate $\x_i$. Although this will degrade the performance of VQ-VAE, as demonstrated in \cref{sec:problem}, the performance of LVD-learned PC can surpass the teacher model. And the primary goal of LVD is to find an initial set of parameters that can be optimized by the EM algorithm to good local optima. We will proceed to show this phenomenon in \cref{sec:lvd-gap-exp}.


\section{Experiments}
\label{sec:exp}

This section first empirically verify the theoretical findings in \cref{sec:problem} (Sec.~\ref{sec:lvd-gap-exp}). We then move on to evaluate our method on image modeling benchmarks (Sec.~\ref{sec:img-exp}).

\subsection{Analyzing Performance Gaps in LVD}
\label{sec:lvd-gap-exp}

We empirically investigate the finding in \cref{sec:problem} that the log-likelihood of the student PC can surpass the ELBO of the teacher DGM. Specifically, we consider an instantiation of the LVD pipeline and empirically compute the three ELBOs shown in \cref{fig:elbo-gap}. 
For ease of computation, we use VQ-VAE as the teacher model and one-shot K-means discretization strategy to train a PC on Imagenet32. The resulting ELBO of the teacher DGM is $-2493$ (Fig.~\ref{fig:elbo-gap} top-right), while the LVD objective is $-2499$ (Fig.~\ref{fig:elbo-gap} bottom). Therefore, the LVD gap is $5$, which matches the extreme case mentioned in \cref{sec:progressive-grow} (\ie the student almost perfectly simulates the teacher). Hence the PC becomes a tractable instantiation of the teacher DGM. Thanks to PC's traceability, we are able to close the variational gap for free and obtain a PC with log-likelihood $-2317$, leading to a student PC better than the teacher DGM.

\subsection{Image Modeling Benchmarks}
\label{sec:img-exp}

We evaluate the proposed algorithmic improvements to the LVD pipeline on three natural image benchmarks: CIFAR \citep{krizhevsky2009learning} and two down-sampled ImageNet (ImageNet32 and ImageNet64) \citep{deng2009imagenet}. 

\textbf{Baselines.}\quad We compare the proposed method, termed LVD with Progressive Growing (LVD-PG) against four TPM baselines: LVD \citep{liu2021tractable}, Hidden Chow-Liu Tree (HCLT) \citep{liu2021tractable}, Random Sum-Product Network (RAT-SPN) \citep{peharz2020random}, and Einsum Network (EiNet). These baselines cover most of the recent endeavors on scaling up and improving the expressiveness of TPMs. Moreover, to evaluate the performance gap with less tractable DGMs, we further compare LVD-PG with the following flow-based models and variational autoencoders: Glow \citep{kingma2018glow}, RealNVP \citep{dinh2016density}, BIVA \citep{maaloe2019biva} and Variational Diffusion Models (VDM) \citep{kingma2021variational}. Implementation details of the baselines can be found in \cref{appx:baseline-details}.


\begin{figure}
    \centering
    \includegraphics[width=\columnwidth]{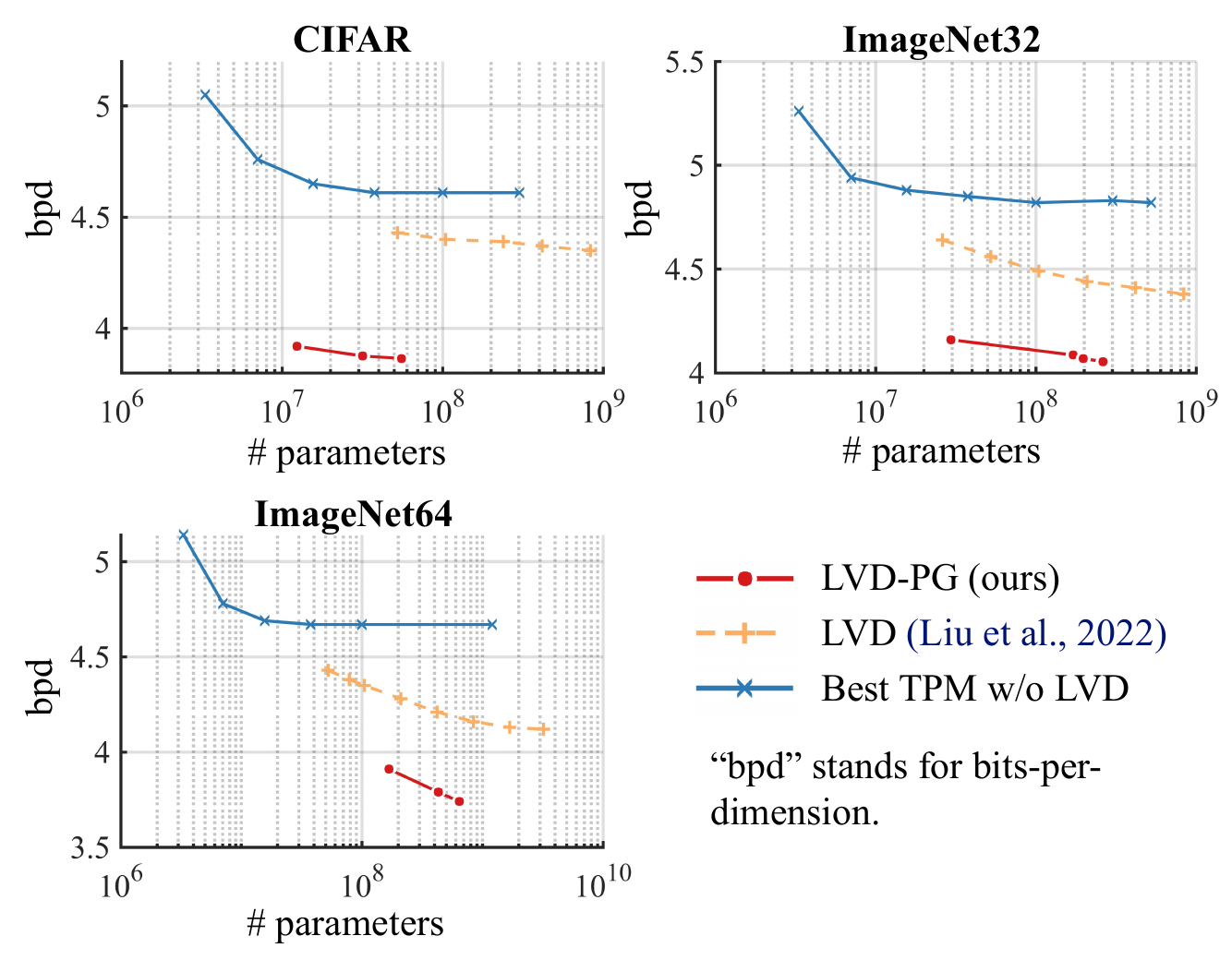}
    \vspace{-2.6em}
    \caption{Comparison of image modeling performance. LVD-PG outperforms all baselines with similar sizes by a large margin.}
    \label{fig:main-results}
    \vspace{-1.2em}
\end{figure}

\textbf{Empirical insights.}\quad We start by comparing our performance with other TPM models. As shown in \cref{fig:main-results}, for all benchmarks, LVD-PG consistently outperforms the previous approaches by a large margin. In particular, on CIFAR, a $\sim\!$ 12M LVD-PG model is much better than a $\sim\!$ 800M PC trained by LVD; on ImageNet32, a $\sim\!$ 20M PC trained by LVD-PG also obtains significant performance gain compared to a $\sim\!$ 800M PC trained by original LVD. This indicates that proper design choices can further exploit LVD's potential to train expressive yet compact PCs, thus significantly boosting the performances of large PCs.

Next, we compare the performance achieved by LVD-PG with the three adopted DGM baselines. Notably, as demonstrated in \cref{tab:img-main-results}, our approach enables PCs to outperform all DGMs except the SoTA VDM on ImageNet64, and on ImageNet32, LVD-PG is only inferior to BIVA with a 0.1 bpd gap and VDM with a 0.34 bpd gap. 

\textbf{Ablation studies.}\quad 
To evaluate the effect of the progressive growing algorithm proposed in \cref{sec:progressive-grow} and the image-data-specific modifications (such as using an ``independent decoder'' in VQ-VAE) elaborated in \cref{sec:image-modeling}, we do an ablation analysis by training two other PCs without either component, respectively. Both PCs have similar model sizes as the SoTA PC trained with LVD-PG on ImageNet32 ($\sim\!$ 260M parameters). Specifically, compared to the SoTA LVD-learned PC with $4.06$ bpd, the LVD-learned PC without progressive growing only achieves $4.12$ bpd, while the performance of the LVD-learned PC with convolutional decoder degrades to $4.18$ bpd.

\vspace{-0.4em}
\section{Related work}
\label{sec:related}

There have been significant recent efforts to scale up and improve the expressiveness of PCs. Many works focus on constructing expressive yet compact initial PC structures \citep{rahman2014cutset,adel2015learning,rooshenas2014learning}, while others aim for an iterative structure learning process that gradually increases model capacity \citep{di2021random,dang2020strudel,liang2017learning}. These methods have led to significant performance gains on various density estimation datasets such as MNIST-family datasets.

However, improving the PC structure alone does not seem to offer too much performance gain on real-world high-dimensional datasets such as natural images and text. Towards solving this problem, there have been many recent endeavors to explore different ways of combining PCs with neural networks (NNs) to obtain tractable while expressive hybrid models. For example, Conditional SPNs \citep{shao2022conditional} harness the expressive power of NNs to learn expressive conditional density estimators; HyperSPNs \citep{shih2021hyperspns} use NNs to regularize the parameters of PCs; \citet{correia2022continuous} learn continuous mixtures of PCs with the help of continuous latent-space models represented by NNs.

A key to the above successes in scaling up PCs is the development of computation frameworks and easy-to-use libraries that make training large-scale PCs highly efficient. Specifically, EiNet \citep{peharz2020einsum} and SPFlow \citep{molina2019spflow} leverage well-developed deep learning packages such as PyTorch \citep{paszke2019pytorch} to implement various inference and parameter learning procedures, and Juice.jl \citep{dang2021juice} implement custom kernels to better handle sparse PCs.

\vspace{-0.4em}
\section{Conclusion}

This paper aims to demystify the latent variable distillation process from intractable Deep Generative Models (DGMs) to tractable Probabilistic Circuits (PCs). We discover both theoretical and empirical evidence that the performance of the student PC can exceed that of the teacher DGM, where the performance gain originates from the tractability of PCs that closes a variational gap ``for free''. Following this variational interpretation of the distillation technique, we further propose algorithmic improvements that lead to significant performance gain over SoTA TPMs. It also outperforms several intractable DGM baselines.

\bibliography{ref}
\bibliographystyle{icml2023}

\newpage
\appendix
\onecolumn

\section{Equivalence Between the Two LVD Formulations}
\label{appx:equiv-lvd}

Consider a sample $\x$. Suppose its LV assignment $\z$ is generated by $\q_{\phi}(\z \given \x)$. Then \cref{eq:lvd-objective} can be written as:
    \begin{align*}
        \mathbb{E}_{\z \sim q_{\phi}(\z \given \x)} [\log p_{pc}(\x,\z)] & = \mathbb{E}_{\z \sim q_{\phi}(\z \given \x)} \left [ \log \left (p_{pc}(\x\given\z)p_{pc}(z) \right ) \right ] , \\
        & = \mathbb{E}_{\z \sim q_{\phi}(\z \given \x)} [\log p_{pc}(\x\given\z)] + \mathbb{E}_{\z \sim q_{\phi}(\z \given \x)}\left [ \log\left ( \frac{p_{pc}(\z)}{q_{\phi}(\z \given \x)}q_{\phi}(\z \given \x) \right ) \right ],\\
        &=\mathbb{E}_{\z \sim q_{\phi}(\z \given \x)} [\log p_{pc}(\x\given\z)]
        - D_{\mathrm{KL}}\left(q_{\phi}(\z \given \x) \| p_{pc}(\z)\right) + \mathbb{E}_{\z \sim q_{\phi}(\z \given \x)} [\log q_{\phi}(\z \given \x)].
    \end{align*}
The first two terms of the last equation are the LVD objective shown in \cref{eq:lvd-vi}. Since the last term is independent with the PC, we conclude that \cref{eq:lvd-objective,eq:lvd-vi} are equivalent up to a constant factor independent with the PC.

\section{Details of the Progressive Growing Algorithm}
\label{appx:prog-grow}

\textbf{Training PC with Labeled Dataset.}\quad For the cluster-conditioned distribution, we adopt multi-head HCLTs with hidden size 16 and run mini-batch EM optimization with batch size 256. The learning rate anneals linearly from 0.1 to 0.01 for 50 epochs.

\textbf{Selecting Clusters to Grow.}\quad The cluster set $\calI$ selected to grow initiates with an empty set, then we choose the cluster with the smallest $\mathrm{LL}$ and push all samples belonging to this cluster into $\calI$ successively until its capacity reaches a certain threshold. In our experiments, the threshold is fixed to be 40$\%$ of the total number of samples.

\textbf{Additional Details.}\quad Given $\data \!:=\! \{(\x, \h) \!:\! \x \!\in\! \data_{\mathrm{train}}, \h \!=\! g_{\phi}(\x)\}$, we first use K-means to pre-cluster the training samples into $N_1$ outer clusters based on their continuous neural embeddings. Then we apply the progressive growing algorithm to grow each outer cluster up to $N_2$ inner clusters, which initiates with a single-head HCLT. Specifically, when $N_1$ equals one, the pipeline is equivalent to growing clusters from scratch, and the smaller total cluster number $N_1\times N_2$ typically corresponds to smaller PCs. Empirically we vary $N_1$ from 50 to 400 and adjust $N_2$ from 20 to 3 accordingly. On the three image benchmarks: Imagenet32, Imagenet64 and CIFAR10, the ($N_1$,$N_2$) adopted by our SoTA LVD-learned PC are (400,4), (320,4), and (100,4), respectively.

\section{Implementation Details of the Baselines}
\label{appx:baseline-details}

To ensure a fair comparison, we implement HCLT and RAT-SPN with the Julia package Juice.jl \citep{dang2021juice} and adopt the original PyTorch implementation of EiNet. For all TPMs, we train a series of models with their number of parameters ranging from $\sim$1M to $\sim$100M and tune hyperparameters accordingly. Finally, we choose the best model among these TPM baselines as the Best TPM w/o LVD. We also report the best performance of each TPM in the following table.

\begin{table}[H]
\centering
\scalebox{1.0}{
\begin{tabular}{lccc}
    \toprule
    Dataset & HCLT & EiNet & RAT-SPN \\
    \midrule
    ImageNet32 & 4.82 & 5.63 & 6.90 \\
    ImageNet64 & 4.67 & 5.69 & 6.82 \\
    CIFAR & 4.61 & 5.81 & 6.95 \\
    \bottomrule
\end{tabular}
}
\vspace{-1.0em}
\end{table}

\end{document}